\documentclass[10pt,twocolumn,letterpaper]{article}

\usepackage[dvipsnames,table]{xcolor}
\usepackage{multirow}
\usepackage{graphicx}
\usepackage{wrapfig,lipsum,booktabs}
\usepackage{lmodern}
\usepackage{ulem}
\usepackage{caption}

\usepackage{iccv}              

\definecolor{iccvblue}{rgb}{0.21,0.49,0.74}
\usepackage[pagebackref,breaklinks,colorlinks,allcolors=iccvblue]{hyperref}

\usepackage[utf8]{inputenc} 
\usepackage[T1]{fontenc}    
\usepackage{hyperref}       
\usepackage{url}            
\usepackage{booktabs}       
\usepackage{amsfonts}       
\usepackage{nicefrac}       
\usepackage{microtype}      
\usepackage{xcolor}         
\usepackage{makecell}
\usepackage{bm}

\usepackage{amsmath,tikz}
\usepackage{amssymb}
\usepackage{mathtools}
\usepackage{amsthm}
\usepackage{arydshln}
\usepackage[accsupp]{axessibility}

\def\eg{\emph{e.g.}} 
\def\ie{\emph{i.e.}} 
 
\def\etc{\emph{etc.}} \def\vs{\emph{vs.}}
 
\def\etal{\emph{et al.}}

\definecolor{mygray}{gray}{.9}
\definecolor{ggray}{RGB}{127,127,127}
\definecolor{reda}{RGB}{192,0,0}
\definecolor{redb}{RGB}{217,148,143}
\definecolor{myyellow}{RGB}{190,144,0}
\definecolor{mygreen}{RGB}{80,100,40}
\definecolor{myblue}{RGB}{30,90,100}

\makeatletter
\newcommand{\thickhline}{%
    \noalign {\ifnum 0=`}\fi \hrule height 1pt
    \futurelet \reserved@a \@xhline
}

\newcommand{\pub}[1]{\color{gray}{\scriptsize{[{#1}]}}}

\makeatletter
\renewcommand{\maketag@@@}[1]{\hbox{\m@th\normalsize\normalfont#1}}%
\makeatother

\title{A Conditional Probability Framework for Compositional Zero-shot Learning}

\author{Peng Wu\textsuperscript{1}\thanks{Equal Contribution.},~ Qiuxia Lai\textsuperscript{2}$^*$,~ Hao Fang\textsuperscript{1},~ Guo-Sen Xie\textsuperscript{3},~ Yilong Yin\textsuperscript{1},~ Xiankai Lu\textsuperscript{1}\thanks{Corresponding author: \textit{Xiankai Lu}.},~ Wenguan Wang\textsuperscript{4,5}
\\ \small{\textsuperscript{1}Shandong University, \textsuperscript{2} Communication University of China, \textsuperscript{3}Nanjing University of Science and Technology, }\\ 
\small{\textsuperscript{4}Zhejiang University, \textsuperscript{5}National Key Laboratory of Human-Machine Hybrid Augmented Intelligence, Xi'an Jiaotong University}}

\begin{document}
\maketitle

\begin{abstract}

Compositional Zero-Shot Learning (CZSL) aims to recognize unseen combinations of known objects and attributes by leveraging knowledge from previously seen compositions. Traditional approaches primarily focus on disentangling attributes and objects, treating them as independent entities during learning. However, this assumption overlooks the semantic constraints and contextual dependencies inside a composition. For example, certain attributes naturally pair with specific objects (\eg, ``striped'' applies to ``zebra'' or ``shirts'' but not ``sky'' or ``water''), while the same attribute can manifest differently depending on context (\eg, ``young'' in ``young tree'' \vs ``young dog''). Thus, capturing attribute-object interdependence remains a fundamental yet long-ignored challenge in CZSL.
In this paper, we adopt a Conditional Probability Framework (CPF) to explicitly model attribute-object dependencies. We decompose the probability of a composition into two components: the likelihood of an object and the conditional likelihood of its attribute. To enhance object feature learning, we incorporate textual descriptors to highlight semantically relevant image regions. These enhanced object features then guide attribute learning through a cross-attention mechanism, ensuring better contextual alignment. By jointly optimizing object likelihood and conditional attribute likelihood, our method effectively captures compositional dependencies and generalizes well to unseen compositions. Extensive experiments on multiple CZSL benchmarks demonstrate the superiority of our approach. Code is available at \href{https://github.com/Pieux0/CPF}{here}. 
\end{abstract}

\section{Introduction}
\label{introduction}

\begin{figure*}[t]
    \centering
    \includegraphics[width=0.98\linewidth]{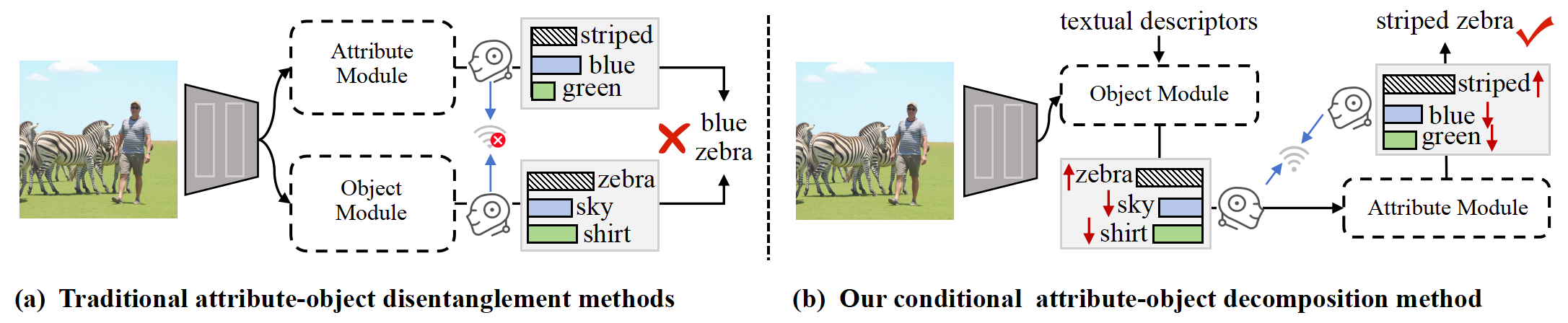}
    \caption{(a) Traditional attribute-object disentanglement methods~\cite{yang2022decomposable,hu2023leveraging,saini2022disentangling,hao2023learning,li2023compositional,dat2023homoe} decompose attributes and objects through separate modules, which fail to capture the inherent attribute-object dependencies. (b) In contrast, we propose a conditional attribute-object decomposition method to model compositional interdependence while incorporating semantic constraints and contextual dependencies.}
    \label{fig:motivation}
    \vskip -0.2in
\end{figure*}

Compositional Zero-Shot Learning (CZSL) is a subfield of zero-shot learning (ZSL) that focuses on recognizing unseen compositions of known objects and attributes by leveraging knowledge from previously observed compositions. 
Most existing CZSL methods assume that attributes and objects are independent and focus on disentangling their representation learning. 
Some approaches~\cite{yang2022decomposable,hu2023leveraging,xu2021relation,karthik2022kg,jiang2024revealing,ruis2021independent,jing2024retrieval} achieve this by processing object and attribute features through separate and independent modules (Fig.~\ref{fig:motivation} (a)). 
Others design complex attention mechanisms as compositional disentanglers, leveraging self-attention~\cite{li2023distilled,liu2023simple} or cross-attention~\cite{saini2022disentangling,hao2023learning,lu2023decomposed,jiang2024mrsp} to learn disentangled object and attribute embeddings. 
However, these methods overlook the semantic constraints and contextual dependencies inherent in attribute-object compositions. 
Semantic constraints dictate that certain attributes naturally pair with specific objects, \eg, ``striped'' typically describes ``zebra'' or ``shirts'' but not ``sky'' or ``water''. 
Contextual dependencies, on the other hand, mean that the visual manifestation of an attribute depends on the object it modifies, \eg, ``young'' appears differently in ``young tree'' \vs ``young dog''.
Fig.\ref{fig:motivation} (a) illustrates the limitations of treating attributes and objects independently. 
When attributes and objects are disentangled, the model assigns similar scores to ``blue'' and ``striped'' in the attribute module based on the image, which can cause erroneous predictions for unseen compositions. 
This issue stems from the fact that an image may contain multiple attributes (e.g., ``blue'', ``striped'', ``green'', \etc), making it challenging to predict the correct attribute in an unseen composition without object information in a fully disentangled manner~\cite{gasser1998learning,misra2017red,nagarajan2018attributes}. 

Recent works have attempted to capture attribute-object contextualization by leveraging object features to generate element-wise attention maps for refining attribute features~\cite{kim2023hierarchical} or by learning module parameters for the attribute learner based on object priors~\cite{wang2023learning}. While these methods address contextual dependency learning to some extent, they remain ineffective in modeling semantic constraints. How to effectively capture the interdependence between attributes and objects remains an open challenge in CZSL.

From a probabilistic perspective~\cite{wang2023learning,kim2023hierarchical,yang2022decomposable}, the likelihood of the composition $c\!=\!(o, a)$ given an image $\bm{x}$ can be decomposed as: $p(o, a|\bm{x})\!=\!p(o|\bm{x})p(a|o, \bm{x})$. Here, $p(o|\bm{x})$ denotes the likelihood of the object given the image, and $p(a|o, \bm{x})$ denotes the likelihood of the attribute conditioned on both the object and the image. A more effective approach to composition learning can be achieved by jointly optimizing these two likelihoods.

Based on this insight, in this paper, we propose a Conditional Probability Framework (CPF) to model compositional interdependence while incorporating semantic constraints and contextual dependencies. To enhance object feature learning, we integrate textual descriptors to highlight semantically relevant image regions. These enhanced object features then guide attribute learning through a cross-attention mechanism, ensuring better contextual alignment. By jointly optimizing object likelihood and conditional attribute likelihood, our method effectively captures compositional dependencies and generalizes well to unseen compositions.

In summary, our contributions are three-fold:
\begin{itemize}[leftmargin=*]
    \item We propose a Conditional Probability Framework (CPF) that models attribute-object dependencies by decomposing composition likelihood into object likelihood and conditional attribute likelihood.
    \item To improve object feature learning, we incorporate textual descriptors to guide object feature learning, focusing on semantically relevant image regions for discriminative representations.
    \item We introduce a cross-attention mechanism that conditions attribute learning on the text-enhanced object features, ensuring better contextual alignment and more accurate attribute-object reasoning.
\end{itemize}

Extensive experiments show that our method achieves state-of-the-art results on three CZSL datasets within both \textit{Closed-world} an \textit{Open-world} settings. In the \textit{Closed-world} setting, our method significantly improves performance, achieving a remarkable +\textbf{17.9}\% AUC on UT-Zappos50K~\cite{yu2014fine}, +\textbf{4.6}\% Seen Accuracy and +\textbf{5.5}\% Unseen Accuracy on MIT-States~\cite{isola2015discovering} and +\textbf{8.1}\%  HM  on C-GQA ~\cite{naeem2021learning}. In the \textit{Open-world} setting, our method continues to outperform existing methods across all datasets, with improvements of +\textbf{8.3}\%  AUC  and  +\textbf{6.3}\% HM  on UT-Zappos50k, +\textbf{175}\%  AUC  and +\textbf{69.7}\% HM on MIT-States, +\textbf{47.9}\% AUC and +\textbf{25.0}\% HM on C-GQA.

\section{Related Work}
\label{related_work}
\subsection{Zero-shot Learning}
Traditional zero-shot Learning (ZSL) aims to recognize unseen classes by leveraging semantic information, such as text descriptions~\cite{reed2016learning}, word embeddings~\cite{socher2013zero}, or attributes~\cite{lampert2013attribute}, that describe those classes. 
To improve generalization to unseen classes, later research has explored various knowledge transfer strategies, including out-of-domain detection~\cite{atzmon2019adaptive,ding2021semantic}, graph neural network~\cite{wang2018zero,xie2020region},  meta-learning~\cite{liu2021task,verma2021meta}, dense attention~\cite{huynh2020fine,huynh2020shared}, and data generation~\cite{xian2018feature}. 
More recently, open vocabulary models such as CLIP~\cite{radford2021learning} have been leveraged for ZSL due to their robust embedding capabilities~\cite{wortsman2022robust,novack2023chils}.
Compositional Zero-Shot Learning (CZSL) extends ZSL by recognizing unseen attribute-object compositions (\eg, ``striped shirts''), where attributes and objects are learned from known compositions during training, and serve as a bridge to generalize to unseen compositions during testing.
In this paper, we focus on CZSL.

\subsection{Compositional Zero-shot Learning}
\noindent\textbf{Learning Compositions as Single-Label Entities.} 
Earlier CZSL methods followed the traditional ZSL paradigm, treating attribute-object compositions as single-label entities and learning to generalize directly to unseen composition labels. 
Some approaches focus on defining transformations between attributes and objects to construct compositional representations from their separate embeddings. 
For example, AOP~\cite{nagarajan2018attributes} factorizes a composition into a matrix-vector product, where the object is represented as a vector and the attribute as a transformation matrix.  
Li~\etal~\cite{li2020symmetry,li2021learning} further proposes three transformations for attribute-object composition based on group axioms and symmetry constraints to enhance compositional embedding learning.  
Other methods~\cite{naeem2021learning,mancini2021open,mancini2022learning,anwaar2022leveraging,ruis2021independent,huang2023reference} leverage graph networks to model relationships between attributes and objects, aiming to learn a more flexible and structured compositional representation with improved compatibility between attributes and objects and enhanced generalization to unseen compositions.
However, with only composition-level learning on a limited set of training compositions, these methods struggle to generalize to the vast number of unseen attribute-object combinations.

\noindent\textbf{Learning Compositions via Attribute-Object Disentanglement.} 
To mitigate the limitations of composition-level learning, researchers have explored disentangling attribute and object representations. 
Some methods achieve this by processing attributes and objects separately through dedicated network modules, such as fully connected layers~\cite{jiang2024revealing}, a combination of convolutional and fully connected layers~\cite{hu2023leveraging}, or multi-layer perceptrons~\cite{li2024tsca,xu2021relation}. 
Others design compositional disentanglers based on attention mechanisms, leveraging self-attention~\cite{li2023distilled,liu2023simple} or cross-attention~\cite{saini2022disentangling,hao2023learning,lu2023decomposed} to learn disentangled attribute and object embeddings. However, these methods fail to capture the inherent dependencies between attributes and objects, where the visual appearance of an attribute can vary significantly when composed with different objects, leading to suboptimal recognition accuracy.

\noindent\textbf{Modeling Contextual Dependencies in Attribute-Object Compositions.}
Rather than focusing on disentangled attribute and object embeddings, recent approaches emphasize capturing their contextual relationships. 
For example, CoT~\cite{kim2023hierarchical} models attribute-object interactions by generating element-wise attention maps conditioned on object features to obtain refined attribute representations. 
CANet~\cite{wang2023learning} conditions attribute embeddings on both the recognized object and the input image and use them as prior knowledge to dynamically adjust the parameters of the attribute learner. 
While these methods help mitigate contextual dependency issues, they still struggle to effectively model semantic constraints between the attribute and object. 
In this paper, we propose a Conditional Probability Framework (CPF) to explicitly model attribute-object dependencies with both semantic constraints and contextual dependencies.

\noindent\textbf{Leveraging Vision-Language Models (VLMs) for CZSL.}
Recent studies have explored VLMs such as CLIP~\cite{radford2021learning,wang2025visual} for CZSL by leveraging their strong zero-shot recognition capabilities. These VLMs are pre-trained on web-scale datasets, which enable compositional generalization through various parameter-efficient fine-tuning techniques~\cite{zhou2022learning,gao2024clip,lu2023drpt,wang2024towards}. 
Some methods use learnable prompts~\cite{nayak2022learning,lu2023decomposed,wang2023hierarchical,huang2024troika,bao2024prompting,qu2025learning,wu2025logiczsl}, while others incorporate lightweight adapters~\cite{zheng2024caila,li2024context} for vision-language alignment. 
Our CPF can also be extended to CLIP by leveraging its text embeddings as semantic constraints to enhance object feature learning, demonstrating its adaptability and scalability.

\section{Methodology}
\label{methodology}
In this section, we first revisit CZSL settings and notations (\S\ref{subsec:ps}). Then, we elaborate on the pipeline of our method CPF (\S\ref{subsec:c2mask}).
Finally, we provide the implementation and reproducibility details (\S\ref{subsec:id}).

\subsection{Problem Statement}
\label{subsec:ps}
In CZSL, given an attribute set $\mathcal{A}=\{a_1,a_2,...,a_M\}$ and an object set  $\mathcal{O}=\{o_1,o_2,...,o_N\}$, the composition set $\mathcal{C}=\{c_1,c_2,...,c_{MN}\}$ is formed as $\mathcal{C}=\mathcal{A} \times \mathcal{O}$ where $c=(a,o)$. Following the task setup, the composition set $\mathcal{C}$ is split into a seen class set $\mathcal{C}_s$ and an unseen class set $\mathcal{C}_u$, ensuring that $\mathcal{C}_s \cap \mathcal{C}_u=\emptyset$. 
The training set is given by $\mathcal{T}=\{(\bm{x},c)|\bm{x} \in \mathcal{X}, c \in \mathcal{C}_s\}$, where each RGB image $\bm{x}$ in the image space $\mathcal{X}$ is labeled with a composition label $c$ from the seen class set $C_s$.
The evaluation is conducted under two settings: \textit{Closed-World (CW)} and \textit{Open-World (OW)}. The corresponding test sets are defined as $\mathcal{T}_{test}^{closed}=\{(\bm{x},c) \mid \bm{x} \in \mathcal{X}, c \in \mathcal{C}_{test}^{closed}\}$ and $\mathcal{T}_{test}^{open}=\{(\bm{x},c) \mid \bm{x} \in \mathcal{X}, c \in \mathcal{C}_{test}^{open}\}$, where $\mathcal{C}_{test}^{closed}=\mathcal{C}_s \cup \mathcal{C}_u'$,  $\mathcal{C}_{test}^{open}=\mathcal{C}_s \cup \mathcal{C}_u$, and $\mathcal{C}_u' \subset \mathcal{C}_u$ is a subset of $\mathcal{C}_u$. 
CZSL aims to learn a mapping: $\mathcal{X} \rightarrow \mathcal{C}_{test}^{open/closed}$ to predict compositions in the test set $\mathcal{T}_{test}^{open/closed}$.

\subsection{Conditional Probability Framework}
\label{subsec:c2mask}
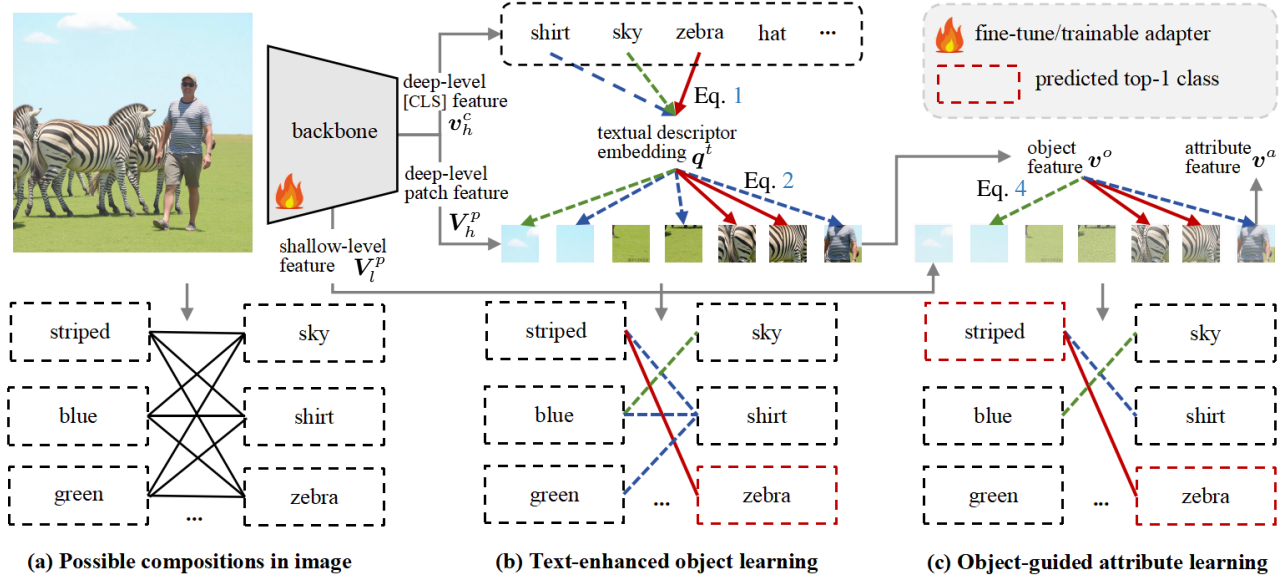
\begin{figure*}
    \centering
    \includegraphics[width=0.99\linewidth]{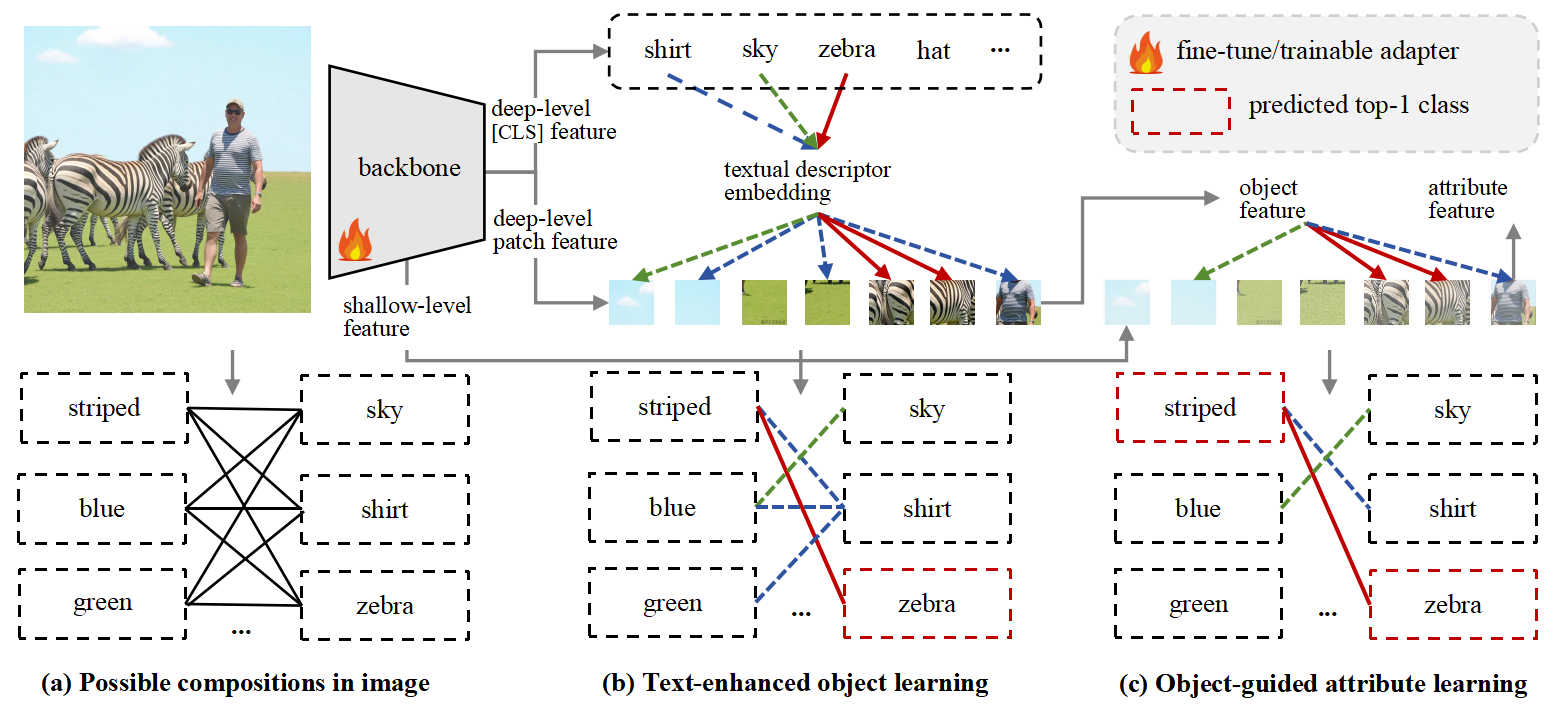}
    \put(-320,172){\small$\bm{v}_h^c$}
    \put(-320,134){\small$\bm{V}_h^p$}
    \put(-356,117){\small$\bm{V}_l^p$}
    \put(-228,159){\small$\bm{q}^t$}
    \put(-227,182){\small Eq.~\ref{eq:q_t}}
    \put(-208,150){\small Eq.~\ref{eq:obj_learning}}
    \put(-78,158){\small$\bm{v}^o$}
    \put(-16,158){\small$\bm{v}^a$}
    \put(-120,147){\small Eq.~\ref{eq:att_learning}}
    \caption{Overall architecture of CPF. (a) Given an image containing certain compositions, our CPF performs decompositions as follows: (b) a \textit{text-enhanced object learning} module, which integrates deep-level visual embeddings with textual embeddings to address semantic constraints and produce enhanced object representations, and (c) an \textit{object-guided attribute learning} module, which captures attribute-object interdependence by learning attribute representations based on text-enhanced object features and shallow-level visual embeddings. }
    \label{fig:architecture}
\end{figure*}

In this paper, we adopt a Conditional Probability Framework (CPF) to explicitly model the interdependence between attributes and objects by incorporating semantic constraints and contextual dependencies, rather than treating them as independent entities. 
As shown in Fig.~\ref{fig:architecture}, our CPF consists of a visual backbone and two key modules: (i) a \textit{text-enhanced object learning} module, which integrates deep-level visual embeddings with textual embeddings to address semantic constraints and produce enhanced object representations, and (ii) an \textit{object-guided attribute learning} module, which captures attribute-object interdependence by learning attribute representations based on text-enhanced object features and shallow-level visual embeddings. 
To ensure alignment between visual and textual features, an additional cross-entropy loss is introduced.
Details are provided in the following. 
Formally, let $[\bm{v}^c_h,\bm{V}_h^p] \in \mathbb{R}^{(1+HW) \times D}$ and $[\bm{v}^c_l,\bm{V}_l^p] \in \mathbb{R}^{(1+HW) \times D}$ denote the deep-level feature and shallow-level feature of image $\bm{x}$ extracted by the visual backbone, respectively.

\noindent\textbf{Text-enhanced Object Learning.} Let the object textual embeddings are represented as $\bm{W}^o=[\bm{w}^o_1,\cdots,\bm{w}^o_N]\in \mathbb{R}^{N \times d}$. 
The text-enhanced object learning module first constructs a textual descriptor embedding $\bm{q}^t\in  \mathbb{R}^{1 \times d}$ by fusing the corresponding object textual embeddings:
\begin{equation}
\begin{aligned}
    \bm{q}^t = \text{softmax}\big(\frac{f_{v\rightarrow t}^o(\bm{v}_h^c) (\bm{W}^o)^{\top}}{\sqrt{d}}\big) \bm{W}^o,
\label{eq:q_t}
\end{aligned}
\end{equation}
where $f_{v\rightarrow t}^o$ is a function that projects visual features into the joint semantic space for text-visual alignment. The textual descriptor embedding $\bm{q}^t$ is then used to enhance semantically relevant image regions by computing its similarity with the set of patch tokens $\bm{V}_h^p$. The resulted attention weights are applied to the image patches, and the refined visual embedding is added to the deep-level class token $\bm{v}_h^c$, yielding the text-enhanced object feature $\bm{v}^o \in \mathbb{R}^{1 \times D}$:
\begin{equation}
\begin{aligned}
    \bm{v}^o = \bm{v}_h^c + \text{softmax}\big(\frac{\bm{q}^t f_{v \rightarrow t}^o(\bm{V}_h^p)^\top}{\sqrt{d}}\big) \bm{V}_h^p .
    \label{eq:obj_learning}
\end{aligned}
\end{equation}
To ensure accurate object classification, we apply a cross-entropy loss $\mathcal{L}_{obj}$ using the text-enhanced object feature $\bm{v}^o$:
\begin{equation}
    \begin{aligned}
         \mathcal{L}_{obj}=\frac{1}{|\mathcal{T}|}\sum\nolimits_{k=1}^{|\mathcal{T}|}-\log p(o|\bm{x}_k), \\  p(o_j|\bm{x}_k)=\frac{\exp(f_{v \rightarrow t}^o(\bm{v}^o_k)\cdot \bm{w}^o_j)}{\sum_{n=1}^N\exp(f_{v \rightarrow t}^o(\bm{v}^o_k)\cdot \bm{w}^o_n)},
    \end{aligned}
    \label{eq:loss_o}
\end{equation}
where $\bm{w}^o_j\in\bm{W}^o$ serves as the weight vector of linear classifier corresponding to object class $o_j$, $k$ indexes the training sample, and $j$ denotes the $j$-th object class. 
Besides object classification, the text-enhanced object feature $\bm{v}^o$ further contributes to guiding attribute learning, as discussed in the following section.

\noindent\textbf{Object-guided Attribute Learning.} 
Let the attribute textual embeddings be represented as $\bm{W}^a=[w^a_1,\cdots,w^a_M]\in \mathbb{R}^{M \times d}$. 
This module explicitly captures attribute-object interdependence through a cross-attention mechanism, where the enhanced object embedding $\bm{v}^o$ attends to the shallow-level patch embeddings $\bm{V}_l^p$: 
\begin{equation}
      \bm{v}^a = \text{softmax}\big(\frac{\bm{v}^o (\bm{V}_l^p)^\top}{\sqrt{D}}\big)\bm{V}_l^p.
    \label{eq:att_learning}
\end{equation}
By computing similarity scores between $\bm{v}^o$ and $\bm{V}_l^p$ followed by a softmax operation, the module assigns higher weights to the most relevant image patches. The resulting weighted sum of patch embeddings forms the attribute representation $\bm{v}^a$, which effectively captures attribute-object interdependence. 

The object-guided attribute learning is achieved through a cross-entropy loss $\mathcal{L}_{att}$ with the object-guided attribute visual feature $\bm{v}^a$:
\begin{equation}
    \begin{aligned}
        \mathcal{L}_{att}=\frac{1}{|\mathcal{T}|}\sum\nolimits_{k=1}^{|\mathcal{T}|}-\log p(a|\bm{x}_k,\bm{v}_k^o),\\
        p(a_i|\bm{x}_k,\bm{v}_k^o)=\frac{\exp(f^a_{v\rightarrow t}(\bm{v}^a_k)\cdot\bm{w}_i^a)}{\sum_{m=1}^M\exp(f^a_{v\rightarrow t}(\bm{v}^a_k)\cdot\bm{w}_m^a)},
    \end{aligned}
    \label{eq:loss_a}
\end{equation}
where $\bm{w}^a_i \in \bm{W}^a$ represents the weight vector of the classifier associated with attribute class $a_i$. The function $f^a_{v\rightarrow t}$ projects the object-guided attribute visual feature $\bm{v}^a_k$ into the joint semantic space for alignment with textual embeddings.  
In this way, the object-guided attribute learning module effectively captures attribute-object dependencies, enhancing compositional generalization.

\noindent\textbf{Composition Matching.} Besides optimizing object and attribute decomposition process, CPF further aligns the compositional visual feature $\bm{v}^c = f^v_c([\bm{v}^a, \bm{v}^o])$ with the compositional textual feature $\bm{w}^c=f^t_c([\bm{w}^a,\bm{w}^o])$ using an additional cross-entropy loss:
\begin{equation}
\begin{aligned}
     \mathcal{L}_{com}=\frac{1}{|\mathcal{T}|}\sum\nolimits_{k=1}^{|\mathcal{T}|}-\log p(c|\bm{x}_k), \\
          p(c_{i,j}|\bm{x}_k) = \frac{\exp(\bm{v}_k^c\cdot \bm{w}_{i,j}^c)}{\sum_{m=1}^M\sum_{n=1}^{N}\exp(\bm{v}_k^c\cdot \bm{w}_{m,n}^c)}.
    \label{eq:loss_c}
\end{aligned}
\end{equation}

\noindent\textbf{Training and Inference.} CPF is jointly optimized by the object classification loss (\ie, $\mathcal{L}_{obj}$), attribute classification loss (\ie, $\mathcal{L}_{att}$) and composition classification loss (\ie, $\mathcal{L}_{com}$):
\begin{equation}
 \mathcal{L}=\mathcal{L}_{com}+\alpha_1 \mathcal{L}_{att}+\alpha_2 \mathcal{L}_{obj},
    \label{eq:loss}
\end{equation}
where $\alpha_1$, $\alpha_2$ are weights that balance the three loss items.

At inference, CPF predicts the composition class $\hat{c}$ from test image $\bm{x}$ by aggregating scores from composition $p(c_{i,j}|\bm{x})$, attribute $p(a_i|\bm{x}, \bm{v}^o)$, and object $p(o_j|\bm{x})$ predictions, using an additive formulation to avoid the multiplicative approach's probability vanishing issue:
\begin{equation}
     \hat{c}=\mathop{\arg \max}\limits_{c_{i,j} \in \mathcal{C}_{test}}  p(c_{i,j}|\bm{x}) +  p(a_i|\bm{x},\bm{v}^o) + p(o_j|\bm{x}).
   \label{eq:pre}
\end{equation}
CPF offers several key merits: \textbf{First}, it comprehensively models attribute-object interdependence. By leveraging text-enhanced object features to guide attribute learning, CPF enforces semantic constraints and contextual dependencies, ensuring more consistent attribute-object predictions. \textbf{Second}, it enhances scalability. CPF can be seamlessly integrated into other CZSL methods via cross-attention, requiring minimal additional trainable parameters.

\subsection{Implementation Details}
\label{subsec:id}
\textbf{Network Architecture.} CPF utilizes a fine-tuned ViT-B model~\cite{dosovitskiy2020image} or a ViT-L/14 in CLIP, as the visual backbone $f^b$. The output of the last block is used as the deep-level visual embedding while the output of $3^{th}$, $6^{th}$ and $9^{th}$ blocks ($6^{th}$, $12^{th}$ and $18^{th}$ blocks for CLIP) are used as shallow-level visual embeddings. Shallow-level features are fused via concatenation and processed through a linear layer. Each embedding consists of a class token $\bm{v}^c_h$ and 196 (256 for CLIP) patch tokens $\bm{V}_h^p$ which are all embedded into 768 (1024 for CLIP) dimensions (\ie, $D\!=\!768$ in Eq.~\ref{eq:att_learning}). To ensure a fair comparison with prior methods, CPF employs GloVe~\cite{pennington2014glove} (or text encoder of CLIP) to encode textual embedding $\bm{W}^a$ and $\bm{W}^o$ for attributes and objects. These textual embeddings are frozen in Glove but remain trainable in CLIP. 
Specifically, the text embedding has 300 (1024 for CLIP) dimensions (\ie, $d\!=\!300$ in Eq.~\ref{eq:q_t} and Eq.~\ref{eq:obj_learning}). The projection function $f^o_{v\rightarrow t}$ and $f^a_{v\rightarrow t}$ are implemented with fully-connected layers.  

\noindent\textbf{Training.} CPF is trained for 10 epochs with Adam optimizer~\cite{kingma2014adam} for all datasets. For ViT-B, the learning rate is set as 1e-4 and decayed by a factor of 0.1 while the learning rate is set as 3.15$\times$1e-6 and decayed by a factor of 1e-5 for CLIP. All loss functions are implemented by cross-entropy loss with the same temperature parameter $\tau\!=\!0.05$. The loss weights $\alpha_1$ and $\alpha_2$ are set to 0.6 and 0.4, respectively (Ablation study can be found in supplementary materials). 

\noindent\textbf{Inference.} We use one input image scale with a shorter side of 224 during inference. CPF introduces a parameter-free token-level attention mechanism, achieving greater efficiency than previous approaches without compromising performance. Our CPF (ViT-B) achieves 1457 fps inference speed, comparable to ADE (1445 fps) and CoT (1460 fps).

\section{Experiment}
\label{experiment}

\subsection{Experimental Details}
\label{sec:exp_details}

\begin{table*}[ht]
    \caption{Evaluation results on MIT-States~\cite{isola2015discovering}, UT-Zappos50K~\cite{yu2014fine} and C-GQA~\cite{naeem2021learning} under \textit{CW} setting. See \S\ref{sec:main_results} for details.}
    \label{tab:closed_word}
    \centering
    \resizebox{\textwidth}{!}{
    \begin{tabular}{|rl|c||cccc|cccc|cccc|}
    \thickhline
    \rowcolor{mygray}
    \multicolumn{2}{|c|}{\textit{Closed-world}} & & \multicolumn{4}{c|}{MIT-States} & \multicolumn{4}{c|}{UT-Zappos50K} & 
    \multicolumn{4}{c|}{C-GQA} \\
    \rowcolor{mygray}
    \multicolumn{2}{|c|}{Method} &  \multirow{-2}{*}{Backbone}  & AUC$\uparrow$ & HM$\uparrow$ & Seen$\uparrow$ & Unseen$\uparrow$ & AUC$\uparrow$ & HM$\uparrow$ & Seen$\uparrow$ & Unseen$\uparrow$ & AUC$\uparrow$ & HM$\uparrow$ & Seen$\uparrow$ & Unseen$\uparrow$ \\ 
    \hline
         AoP~\cite{nagarajan2018attributes}\!\!\! &\!\!\!\pub{ECCV2018} & ResNet18 & 1.6 & 9.9 & 14.3 & 17.4 & 25.9	& 40.8 & 59.8 & 54.2 & 0.3 & 2.9 & 11.8 & 3.9 \\
         TMN~\cite{purushwalkam2019task}\!\!\! &\!\!\!\pub{ICCV2019} & ResNet18 & 2.9 & 13 & 20.2 & 20.1 & 29.3 & 45 & 58.7	& 60 & 1.1 & 7.7 & 21.6 & 6.3 \\
         SymNet~\cite{li2020symmetry}\!\!\! &\!\!\!\pub{CVPR2020} & ResNet18 & 3.0 & 16.1 & 24.4	&25.2 &23.4	&40.4 &49.8	&57.4 &2.2	&10.9 &27.0 &10.8 \\
         CompCos~\cite{mancini2021open}\!\!\! &\!\!\!\pub{CVPR2021} &ResNet18 & 4.8 &16.9 &26.9 &24.5 &31.8 &48.1 &58.8 &63.8 &2.9	&12.8 &30.7	&12.2 \\
         CGE~\cite{naeem2021learning}\!\!\! &\!\!\!\pub{CVPR2021} &ResNet18 & 5.1 &17.2 &28.7 &25.3 &26.4 &41.2 &56.8 &63.6 &2.5 &11.9	&27.5 &11.7 \\
         Co-CGE~\cite{mancini2022learning}\!\!\! &\!\!\!\pub{TPAMI2022} & ResNet18& - & - & - & - & 30.8 &44.6 &60.9 &62.6 &3.6	&14.7 &31.6	&14.3 \\
         SCEN~\cite{li2022siamese}\!\!\! &\!\!\!\pub{CVPR2022} &ResNet18 & 5.3	&18.4 &29.9	&25.2 &30.9 &46.7 &\underline{65.7}	&62.9 &3.5	&14.6 &31.7	&13.4  \\
         OADis~\cite{saini2022disentangling}\!\!\! &\!\!\!\pub{CVPR2022} &ResNet18 &  5.9 &18.9	&31.1 &25.6	&32.6 &46.9	&60.7 &\underline{68.8}	&3.8 &14.7 &33.4 &14.3 \\
        
         IVR~\cite{zhang2022learning}\!\!\! &\!\!\!\pub{ECCV2022} & ResNet18& - & - & - & - & 34.3	&49.2 &61.5	&68.1 &2.2	&10.9 &27.3	&10.0 \\
         CAPE~\cite{khan2023learning}\!\!\! &\!\!\!\pub{WACV2023}& ResNet18& 5.8	&19.1 &30.5	&26.2 & - & - & - & - & 4.2	&16.3 &32.9	&15.6   \\
         CANet~\cite{wang2023learning}\!\!\! &\!\!\!\pub{CVPR2023}& ResNet18& 5.4	&17.9 &29.0 &26.2 &33.1	&47.3 &61 &66.3	&3.3 &14.5 &30	&13.2   \\
         CGE~\cite{naeem2021learning}\!\!\! &\!\!\!\pub{CVPR2021} &ViT-B & 9.7 & 24.8 & \underline{39.7} & 31.6 & - &- &-	&- &5.4 &18.5 &38.0 &17.1   \\
          OADis~\cite{saini2022disentangling}\!\!\! &\!\!\!\pub{CVPR2022} &ViT-B & 10.1 & 25.2 & 39.2 & 32.1 & - &-	&-	&- &7.0 &20.1 &38.3 &19.8   \\
         ADE~\cite{hao2023learning}\!\!\! &\!\!\!\pub{CVPR2023} & ViT-B & - & - & - & - & \underline{35.1} &\underline{51.1}	&63	&64.3 &5.2 &18.0 &35 &17.7   \\
         CoT~\cite{kim2023hierarchical}\!\!\! &\!\!\!\pub{ICCV2023} & ViT-B & \underline{10.5} &\underline{25.8}	&39.5 &\underline{33.0} & - & - & - & - & \underline{7.4} &\underline{22.1}	&\underline{39.2} &\underline{22.7} \\  
         \hdashline
         \textbf{CPF} \!\!\! &\!\!\!\!\! \textbf{(Ours)}  & ViT-B& \textbf{11.2} &\textbf{26.8} &\textbf{41.3}	&\textbf{34.8} &\textbf{41.4} &\textbf{55.7} &\textbf{66.4}	&\textbf{71.1} & \textbf{8.2} &\textbf{23.9} &\textbf{39.6}	&\textbf{23.5} \\
    \thickhline
    \end{tabular}}
\end{table*}

\textbf{Datasets.} CPF is evaluated on three  widely-used CZSL benchmarks: UT-Zappos50K~\cite{yu2014fine}, MIT-States~\cite{isola2015discovering}, and C-GQA~\cite{naeem2021learning}. UT-Zappos50K~\cite{yu2014fine} includes an extensive collection of shoe types (\eg, Shoes.Heels, Boots.Ankle) and various material properties (\eg, Cotton, Nylon). MIT-States~\cite{isola2015discovering} features 115 attributes (\eg, ancient, broken) and 245 objects (\eg, computer, tree), presenting a substantially broader compositional scope than UT-Zappos50K. C-GQA~\cite{naeem2021learning} is the most extensive CZSL dataset, featuring 453 states, 870 objects, 39,298 images, and more than 9,500 distinct state-object combinations. The split details of the above benchmarks are summarized in supplementary materials.

\noindent\textbf{Metrics.} 
To comprehensively evaluate the effectiveness of CPF, we report four metrics. In particular, Seen Accuracy is calculated for evaluating the performance on seen compositions while Unseen Accuracy is computed for evaluating the classification performance on unseen compositions. With Seen Accuracy as $x-$axis and Unseen Accuracy as $y-$axis, we derive a seen-unseen accuracy curve. We then compute and report the area under the curve (AUC) as well as the best harmonic mean (HM). Following previous literature~\cite {hao2023learning,mancini2021open}, we apply calibration terms to alleviate the bias towards seen compositions for fair comparison.

\noindent\textbf{Evaluation Settings.} 
Following previous approaches~\cite{hao2023learning,lu2023decomposed}, we perform evaluations under both the \textit{CW} and \textit{OW} settings~\cite{mancini2021open,huo2024procc}. The \textit{CW} protocol serves as the standard evaluation framework, considering only a predefined subset of compositions during the testing phase. In contrast, the \textit{OW} setting is designed for a more exhaustive assessment, encompassing all possible composition classes.

\begin{table*}[ht]
    \caption{Evaluation results on MIT-States~\cite{isola2015discovering}, UT-Zappos50K~\cite{yu2014fine} and C-GQA~\cite{naeem2021learning} under \textit{OW} setting. See \S\ref{sec:main_results} for details.}
    \label{tab:open_word}
    \centering
    \resizebox{\textwidth}{!}{
    \begin{tabular}{|rl|c||cccc|cccc|cccc|}
    \thickhline
    \rowcolor{mygray}
    \multicolumn{2}{|c|}{\textit{Open-world}}&  & \multicolumn{4}{c|}{MIT-States} & \multicolumn{4}{c|}{UT-Zappos50K} & 
    \multicolumn{4}{c|}{C-GQA} \\
    \rowcolor{mygray}
   \multicolumn{2}{|c|}{Method} &  \multirow{-2}{*}{Backbone} & AUC$\uparrow$ & HM$\uparrow$ & Seen$\uparrow$ & Unseen$\uparrow$ & AUC$\uparrow$ & HM$\uparrow$ & Seen$\uparrow$ & Unseen$\uparrow$ & AUC$\uparrow$ & HM$\uparrow$ & Seen$\uparrow$ & Unseen$\uparrow$ \\ 
    \hline
         AoP~\cite{nagarajan2018attributes}\!\!\! &\!\!\!\pub{ECCV2018} & ResNet18& 0.7 & 4.7 & 16.6 & 5.7 & 13.7	& 29.4 & 50.9 & 34.2 & - & - & - & - \\
         TMN~\cite{purushwalkam2019task}\!\!\! &\!\!\!\pub{ICCV2019} &ResNet18 & 0.1 & 1.2 & 12.6 & 0.9 & 8.4 & 21.7 & 55.9 & 18.1 & - & - & - & - \\
         SymNet~\cite{li2020symmetry}\!\!\! &\!\!\!\pub{CVPR2020} &ResNet18 & 0.8 & 5.8 & 21.4 &7.0 &18.5	&34.5 &53.3	&44.6 &0.43	&3.3 &26.7 &2.2 \\
         CompCos~\cite{mancini2021open}\!\!\! &\!\!\!\pub{CVPR2021} &ResNet18 & 1.6 &8.9 &25.4 &10.0 &21.3 &36.9 &59.3 &46.8 &0.39	&2.8 &28.4	&1.8 \\
         CGE~\cite{naeem2021learning}\!\!\! &\!\!\!\pub{CVPR2021} &ResNet18 & 1.0 &6.0 &\underline{32.4} &5.1 &23.1 &39.0 &61.7 &47.7 &0.47 &2.9	&32.7 &1.8 \\
         OADis~\cite{saini2022disentangling} \!\!\! &\!\!\!\pub{CVPR2022} &ResNet18 & - &-	&- &-	&25.3 &41.6 &58.7 &53.9	&0.71 &4.2 &33.0 &2.6 \\
         KG-SP~\cite{karthik2022kg}\!\!\! &\!\!\!\pub{CVPR2022} &ResNet18 & 1.3 &7.4 &28.4 &7.5 &26.5 &42.3 &61.8 &52.1 &0.78 &4.7 &31.5 &2.9 \\
         DRANet~\cite{li2023distilled} \!\!\! &\!\!\!\pub{ICCV2023} &ResNet18 & 1.5 & 7.9 & 29.8 & 7.8 & \underline{28.8} &44.0	&\textbf{65.1}	&\underline{54.3} &1.05 &6.0 &31.3 &3.9 \\
         ProCC~\cite{huo2024procc}\!\!\! &\!\!\!\pub{AAAI2024} &ResNet18 & \underline{1.9} & \underline{10.7} & 31.9 & \underline{11.3} & 27.9 &43.8	&\underline{64.8}	&51.5 &0.91 &5.3 &33.2 &3.2 \\
          Co-CGE~\cite{mancini2022learning}\!\!\! &\!\!\!\pub{TPAMI2022}& ViT-B & -& - & - & - & 22.0 &40.3 &57.7 &43.4 & 0.48 &3.3 &31.1 &2.1 \\
        OADis~\cite{saini2022disentangling}\!\!\! &\!\!\!\pub{CVPR2022} & ViT-B & - &-	&- &- & 25.3 &41.6 &58.7 &53.9 & 0.71 &4.2 &33.0 &2.6 \\
           IVR~\cite{zhang2022learning}\!\!\! &\!\!\!\pub{ECCV2022} & ViT-B & -& - & - & - & 25.3 &42.3 &60.7 &50.0 & 0.94 &5.7 &30.6 &4.0 \\
         ADE~\cite{hao2023learning}\!\!\! &\!\!\!\pub{CVPR2023} & ViT-B & - & - & - & - & 27.1 &\underline{44.8} &62.4 &50.7 &\underline{1.42} &\underline{7.6} &\underline{35.1} &\underline{4.8}   \\
        \hdashline
         \textbf{CPF} \!\!\! &\!\!\!\!\! \textbf{(Ours)}& ViT-B & \textbf{4.4} &\textbf{15.1}&\textbf{40.8} &\textbf{14.4} & \textbf{31.2} & \textbf{47.6} & 64.6 & \textbf{56.1} & \textbf{2.10} & \textbf{9.5} & \textbf{38.4} & \textbf{6.8} \\
    \thickhline
    \end{tabular}}
\end{table*}

\subsection{Main Results}
\label{sec:main_results}
In this section, we evaluate and analyze the performance of CPF against state-of-the-art methods across three CZSL datasets (\ie, UT-Zappos50K~\cite{yu2014fine}, MIT-States~\cite{isola2015discovering}, and C-GQA~\cite{naeem2021learning}) under both \textit{CW}  and \textit{OW} settings. The results are reported in Table~\ref{tab:closed_word} and Table~\ref{tab:open_word}. Furthermore, we integrate the proposed CPF into CLIP to assess its effectiveness and scalability. The corresponding experimental results for both settings are detailed in Table~\ref{tab:clip_result}. 

\noindent\textbf{Performance in the \textit{CW} Setting.} As shown in Table~\ref{tab:closed_word}, our proposed CPF method surpasses recent state-of-the-art (SOTA) CZSL approaches~\cite{hao2023learning,wang2023learning,kim2023hierarchical,saini2022disentangling} across all datasets in the \textit{CW} setting. 
Notably, in terms of AUC---the most representative and stable metric for evaluating CZSL model performance~\cite{hao2023learning}---CPF achieves significant improvements: +\textbf{6.7\%} on MIT-States, +\textbf{17.9\%} on UT-Zappos50K, and + \textbf{10.8\%} on C-GQA compared to the SOTA methods. Furthermore, CPF boosts HM to \textbf{26.8} (+\textbf{3.9}\%), \textbf{55.7} (+\textbf{9.0}\%) and \textbf{23.9} (+\textbf{8.1}\%) on MIT-States, UT-Zappos50K and C-GQA. In addition, CPF yields +\textbf{4.0}\%, +\textbf{1.1}\% and +\textbf{1.0}\% Seen Accuracy score gains, as well as +\textbf{5.5}\%, +\textbf{3.3}\% and +\textbf{3.5}\% Unseen Accuracy score gains on MIT-States, UT-Zappos50K and C-GQA. These performance gains can be attributed to CPF's effectiveness in modeling the interdependence between attributes and objects. 

\noindent\textbf{Performance in the \textit{OW} Setting.} Performing classification in the \textit{OW} setting is considerably more challenging, as it requires evaluating all possible attribute-object compositions. Consequently, most CZSL methods experience a significant drop in performance under this setting. To address this challenge, certain methods, such as KG-SP~\cite{karthik2022kg} and DRANet~\cite{li2023distilled}, leverage external knowledge to reduce the number of composition classes. In contrast, CPF still obtains the best performance on almost all evaluation metrics (see Table~\ref{tab:open_word}) without using external knowledge. Specifically, CPF boosts AUC to \textbf{4.4} (+\textbf{175}\%) on MIT-States, \textbf{31.2} (+\textbf{8.3}\%) and \textbf{2.10} (+\textbf{47.9}\%). Beyond AUC, CPF achieves notable improvements in HM, Seen Accuracy, and Unseen Accuracy on all datasets. These performance improvements reinforce our belief that capturing semantic constraints and contextual dependencies in attribute-object compositions is essential for identifying novel combinations, even under the challenging conditions of the \textit{OW} setting.

\noindent\textbf{Performance with the CLIP Backbone.} To further validate the efficacy and scalability of our proposed CPF, we develop a CLIP-based implementation of the CPF model. As summarized in Table~\ref{tab:clip_result}, CPF outperforms state-of-the-art CLIP-based CZSL methods on the most challenging CZSL benchmark (\ie, C-GQA) under both \textit{CW} and \textit{OW} settings.

\begin{table}[ht]
    \caption{Evaluation with CLIP-based CPF. See \S\ref{sec:main_results} for details.}
    \label{tab:clip_result}
    \centering
    \resizebox{0.48\textwidth}{!}{
    \begin{tabular}{|rl|c||cccc|}
    \thickhline
    \rowcolor{mygray}
    & &  & \multicolumn{4}{c|}{C-GQA} \\
    \rowcolor{mygray}
    \multicolumn{2}{|c|}{\multirow{-2}{*}{Method}} & \multirow{-2}{*}{Backbone} & AUC$\uparrow$ & HM$\uparrow$ & Seen$\uparrow$ & Unseen$\uparrow$  \\ 
    \hline
     \multicolumn{7}{|c|}{\textit{Closed-world}} \\ \hline
CoOp~\cite{zhou2022learning}\!\!\! &\!\!\!\pub{IJCV2022} & CLIP   & 4.4 & 17.1 & 20.5 & 26.8 \\
        CSP~\cite{nayak2022learning}\!\!\! &\!\!\!\pub{ICLR2023} & CLIP  & 6.2 & 20.5 & 28.8 & 26.8\\
        DFSP~\cite{lu2023decomposed}\!\!\! &\!\!\!\pub{CVPR2023}& CLIP & 10.5 & 27.1 & 38.2 & 32.0\\
        CDS-CZSL~\cite{li2024context}\!\!\! &\!\!\!\pub{CVPR2024}& CLIP & 11.1 & 28.1 & 38.3 & 34.2\\
        Troika~\cite{huang2024troika}\!\!\! &\!\!\!\pub{CVPR2024}& CLIP  & 12.4 & 29.4 & 41.0 & 35.7\\
        PLID~\cite{bao2024prompting}\!\!\! &\!\!\!\pub{ECCV2024}& CLIP   & 11.0 & 27.9 & 38.8 & 33.0\\
        CAILA~\cite{zheng2024caila} \!\!\! &\!\!\!\pub{WACV2024} & CLIP  & 14.8 & 32.7 & 43.9	& 38.5\\
         C{\footnotesize LUS}P{\footnotesize RO}~\cite{qu2025learning} \!\!\! &\!\!\!\pub{ICLR2025} & CLIP & 14.9 & 32.8 & 44.3 & 37.8 \\
          L{\footnotesize OGI}C{\footnotesize ZSL}~\cite{wu2025logiczsl} \!\!\! &\!\!\!\pub{CVPR2025} & CLIP & \underline{15.3} & \underline{33.3} & \underline{44.4} & \underline{39.4} \\
         \hdashline
        \textbf{CPF} \!\!\! &\!\!\!\!\! \textbf{(Ours)} & CLIP & \textbf{15.4} & \textbf{33.6} & \textbf{44.8}	& \textbf{39.6}\\ \hline
         \multicolumn{7}{|c|}{\textit{Open-world}} \\ \hline
        CoOp~\cite{zhou2022learning}\!\!\! &\!\!\!\pub{IJCV2022} & CLIP & 0.7 & 5.5 & 21.0 & 4.6\\
        CSP~\cite{nayak2022learning}\!\!\! &\!\!\!\pub{ICLR2023} & CLIP   & 1.2 & 6.9 & 28.7 & 5.2\\
        DFSP~\cite{lu2023decomposed}\!\!\! &\!\!\!\pub{CVPR2023} & CLIP  & 2.4 & 10.4 & 38.3 & 7.2 \\
        CDS-CZSL~\cite{li2024context}\!\!\! &\!\!\!\pub{CVPR2024}& CLIP & 2.7 & 11.6 & 37.6 & 8.2 \\
        Troika~\cite{huang2024troika}\!\!\! &\!\!\!\pub{CVPR2024} & CLIP  & 2.7 & 10.9 & 40.8 & 7.9\\
        PLID~\cite{bao2024prompting}\!\!\! &\!\!\!\pub{ECCV2024} & CLIP  & 2.5 & 10.6 & 39.1 & 7.5 \\
        CAILA~\cite{zheng2024caila}\!\!\! &\!\!\!\pub{WACV2024} & CLIP   & 3.1 & 11.5 & \underline{43.9}	& 8.0 \\
        C{\footnotesize LUS}P{\footnotesize RO}~\cite{qu2025learning}\!\!\! &\!\!\!\pub{ICLR2025}& CLIP &3.0 &11.6 & 41.6& 8.3\\
         L{\footnotesize OGI}C{\footnotesize ZSL}~\cite{wu2025logiczsl} \!\!\!
        &\!\!\!\pub{CVPR2025} & CLIP & \underline{3.4} & \underline{12.6} & 43.7 & 9.3 \\
        \hdashline
        \textbf{CPF} \!\!\! &\!\!\!\!\! \textbf{(Ours)} & CLIP & \textbf{3.6} & \textbf{13.0} &	\textbf{44.5}& \textbf{9.3} \\
    \thickhline    \end{tabular}}
\end{table}

\subsection{Ablation Experiments}
\label{sec:dig_exp}
To evaluate our algorithm designs and gain further insights, we carry out comprehensive ablation studies on C-GQA~\cite{naeem2021learning} under both \textit{CW} and \textit{OW} settings.

\noindent\textbf{Key Component Analysis.} We first examine the essential components of CPF in Table~\ref{tab:components}. Here TEO and OGA denote the text-enhanced object learning and object-guided attribute learning. We observe a notable performance decline in both \textit{CW} and \textit{OW} settings when the TEO component is removed. This verifies the efficacy of incorporating textual descriptors into object decomposition process. Additionally, the removal of the OGA component leads to a further degradation in model performance, which confirms the significance of attribute-object interdependence in attribute learning.

\begin{table}[ht]
    \caption{Analysis of essential components on C-GQA~\cite{naeem2021learning}.}
    \centering
    \resizebox{0.47\textwidth}{!}{
    \begin{tabular}{|c|l||cccc|}
    \thickhline
    \rowcolor{mygray}
    & & \multicolumn{4}{c|}{C-GQA} \\
    \rowcolor{mygray}
     \multirow{-2}{*}{Setting} &  \multirow{-2}{*}{Methods} & AUC$\uparrow$ & HM$\uparrow$ & Seen$\uparrow$ & Unseen$\uparrow$ \\ 
    \hline
    \multirow{3}{*}{\textit{Closed-world}} 
     & Full & \textbf{8.2} &	\textbf{23.9}	&\textbf{39.6}	&\textbf{23.5} \\
    & -TEO & 7.6 &22.7	&39.6	&22.0  \\
    & -TEO-OGA & 6.9 &21.4	&37.8 &21.6  \\
    \hline
    \multirow{3}{*}{\textit{Open-world}} 
    & Full & \textbf{2.10}	&\textbf{9.5}	&38.4	&\textbf{6.8}  \\
    & -TEO & 1.79	&8.3	&\textbf{38.6}	&5.6  \\
    & -TEO-OGA & 1.69	&7.9	&38.3	&5.3  \\
    \hline
    \end{tabular}}
\label{tab:components}
\end{table}

\noindent\textbf{Attention Module.} We next investigate the effectiveness of cross-attention design in Table~\ref{tab:attention}. We can find that, replacing the attention module in Eq.~\ref{eq:obj_learning} and Eq.~\ref{eq:att_learning} with a simple averaging operation results in a significant performance drop. This verifies the effectiveness of the cross-attention mechanism in improving contextual alignment. 

\begin{table}[ht]
\centering
    \caption{Analysis of cross-attention design on C-GQA~\cite{naeem2021learning}.}
    \centering
    \resizebox{0.48\textwidth}{!}{
    \begin{tabular}{|c|l||cccc|}
    \thickhline
    \rowcolor{mygray}
    & &  \multicolumn{4}{c|}{C-GQA} \\
    \rowcolor{mygray}
   \multirow{-2}{*}{Setting} & \multirow{-2}{*}{Methods} & AUC$\uparrow$ & HM$\uparrow$ & Seen$\uparrow$ & Unseen$\uparrow$ \\ \hline
    \multirow{4}{*}{\textit{Closed-world}} 
    & average (Eq.~\ref{eq:obj_learning}) & 7.8 & 22.9 & 39.1& 23.0  \\
    & attention (Eq.~\ref{eq:obj_learning}) & 8.2 &	23.9	&39.6	&23.5  \\
    \cdashline{2-6}
    & average (Eq.~\ref{eq:att_learning}) & 7.1& 22.0&37.9 & 21.4  \\
    & attention (Eq.~\ref{eq:att_learning}) & 8.2 &	23.9	&39.6	&23.5  \\
    \hline
    \multirow{4}{*}{\textit{\makecell[c]{Open-world}}} 
    & average (Eq.~\ref{eq:obj_learning}) & 1.91 & 8.5 & 38.6 & 5.9  \\
    & attention (Eq.~\ref{eq:obj_learning}) & 2.10	&9.5 &38.4	&6.8  \\
    \cdashline{2-6}
    & average (Eq.~\ref{eq:att_learning}) & 1.79 & 8.1 & 37.8 & 5.9 \\
    & attention (Eq.~\ref{eq:att_learning}) & 2.10 &9.5	&38.4	&6.8  \\
    \thickhline
    \end{tabular}}
\label{tab:attention}
\end{table}

\noindent\textbf{Visual Embedding Choice.} Table~\ref{tab:features} probes the impact of visual embedding choice for object and attribute decomposition. Following previous methods~\cite{hao2023learning,li2023distilled}, we initially select deep-level visual embeddings for disentangling object and attribute representations. Our model CPF achieves significant improvements (\ie, AUC: 5.2 → \textbf{6.7} and 1.42 → \textbf{1.58}, HM: 18.0 → \textbf{20.8} and 7.6 → \textbf{7.7}) in both \textit{CW} and \textit{OW} settings compared to ADE~\cite{hao2023learning}, which employs the same visual embeddings. This confirms that our proposed CPF is more effective than those approaches that treat attribute and object as independent entities. Moreover, employing both deep-level and shallow-level visual embedding yields notable performance gains over relying solely on deep-level embeddings. This highlights the necessity of fine-grained information for effective attribute learning~\cite{sarafianos2018deep}.

\begin{table}[ht]
\centering
    \caption{Impact of visual embedding choice in attribute and object decomposition learning on C-GQA~\cite{naeem2021learning}.}
    \centering
    \resizebox{0.47\textwidth}{!}{
    \begin{tabular}{|c|l||cccc|}
    \thickhline
    \rowcolor{mygray}
    & &  \multicolumn{4}{c|}{C-GQA} \\
    \rowcolor{mygray}
   \multirow{-2}{*} {Setting} & \multirow{-2}{*}{Methods} & AUC$\uparrow$ & HM$\uparrow$ & Seen$\uparrow$ & Unseen$\uparrow$ \\ \hline
    \multirow{3}{*}{\textit{\makecell[c]{Closed-world}}} 
    & ADE~\cite{hao2023learning} &5.2 &18.0 &35.0 &17.7   \\
    & deep-level & 6.7 & 20.8 & 37.1 & 21.8  \\
    & shallow+deep-level & \textbf{8.2} & \textbf{23.9} & \textbf{39.6} & \textbf{23.5} \\
    \hline
    \multirow{3}{*}{\textit{\makecell[c]{Open-world}}} & ADE~\cite{hao2023learning} &1.42 &7.6 &35.1 &4.8   \\
    & deep-level & 1.58 & 7.7 & 36.5 & 5.4  \\
    & shallow+deep-level & \textbf{2.10} &\textbf{9.5} &\textbf{38.4} &\textbf{6.8} \\
    \thickhline
    \end{tabular}}
\label{tab:features}
\end{table}

\noindent\textbf{Impact of Guidance in Attribute Learning.} We examine the impact of guidance in attribute learning in Eq.~\ref{eq:att_learning}.  As shown in Table~\ref{tab:guided}, we replace object\_visual embedding $\bm{v}^o$ in Eq.~\ref{eq:att_learning} with attribute textual embedding $\bm{W}^a$ for guiding attribute learning. We observe a significant performance drop across key metrics (\eg, AUC: 8.2 $\rightarrow$ 7.6, 2.10 $\rightarrow$ 1.83) in both \textit{CW} and \textit{OW} settings, primarily due to the model's inability to capture the interdependence between attributes and objects. We subsequently leverage the object textual embedding $\bm{W}^o$ as a guiding signal for attribute learning. The results reveal that CPF outperforms methods relying on attribute textual embeddings, yet it remains less effective than approaches utilizing object visual embeddings. This phenomenon occurs because visual embeddings exhibit stronger alignment with attributes, as visual features inherently capture the characteristic properties of attributes, whereas textual embeddings rely on semantic associations derived from object names, frequently failing to accurately represent the visual relationships between objects and attributes.

\begin{table}
\centering
    \caption{\small{Impact of guidance in attribute learning.}}
    \centering
    \renewcommand{\arraystretch}{1.1}
    \resizebox{0.48\textwidth}{!}{
    \begin{tabular}{|c|l||cccc|}
    \thickhline
    \rowcolor{mygray}
    & & \multicolumn{4}{c|}{C-GQA} \\
    \rowcolor{mygray}
    \multirow{-2}{*}{Setting} & \multirow{-2}{*} {Guidance} & AUC$\uparrow$ & HM$\uparrow$ & Seen$\uparrow$ & Unseen$\uparrow$
     \\  \hline
    \multirow{3}{*}{\textit{Closed-world}} 
    & attribute\_text embedding & 7.6 & 22.6 & 38.7 & 22.8  \\
    & object\_text embedding & 7.7 & 22.7 & 39.0& 22.9  \\
    & object\_visual embedding & \textbf{8.2} & \textbf{23.9} & \textbf{39.6} & \textbf{23.5} \\
    \hline
    \multirow{3}{*}{\textit{Open-world}} 
    & attribute\_text embedding & 1.83	&8.1 &\textbf{38.6}	&5.7  \\
    & object\_text embedding & 1.90 &8.6 &38.0 &5.9  \\
    & object\_visual embedding & \textbf{2.10} &\textbf{9.5} &38.4 &\textbf{6.8}  \\
    \thickhline
    \end{tabular}}
\label{tab:guided}
\end{table} 

\begin{figure*}[ht]
    \centering
    \includegraphics[width=1\linewidth]{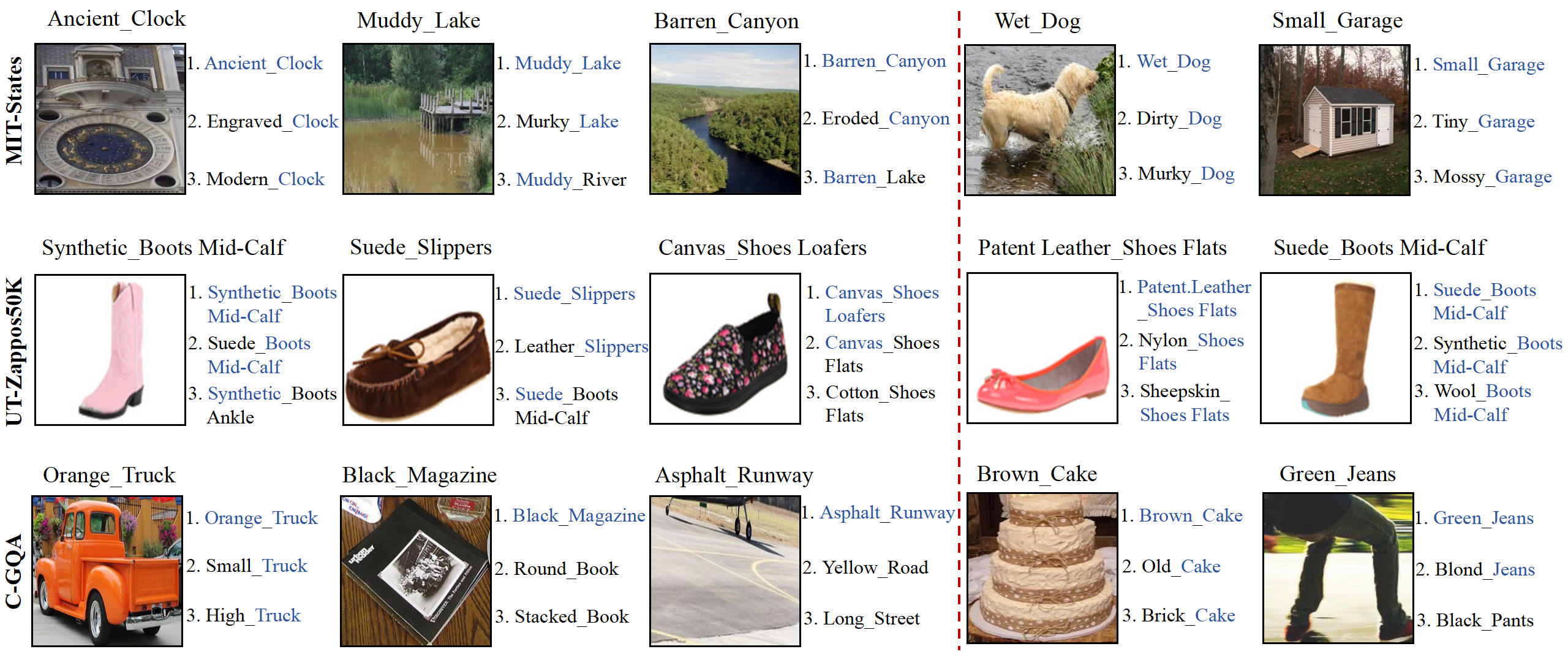}
    \caption{For qualitative results: we demonstrate Top-3 predictions of our proposed CPF model for each sampled instance on UT-Zappos50K~\cite{yu2014fine}, MIT-States~\cite{isola2015discovering}, and C-GQA~\cite{naeem2021learning} under \textit{CW} (left) and \textit{OW} (right) settings. Blue text indicates correct predictions.}
    \label{fig:visualization}
\end{figure*}

\begin{figure}[ht]
    \centering
    \includegraphics[width=0.98\linewidth]{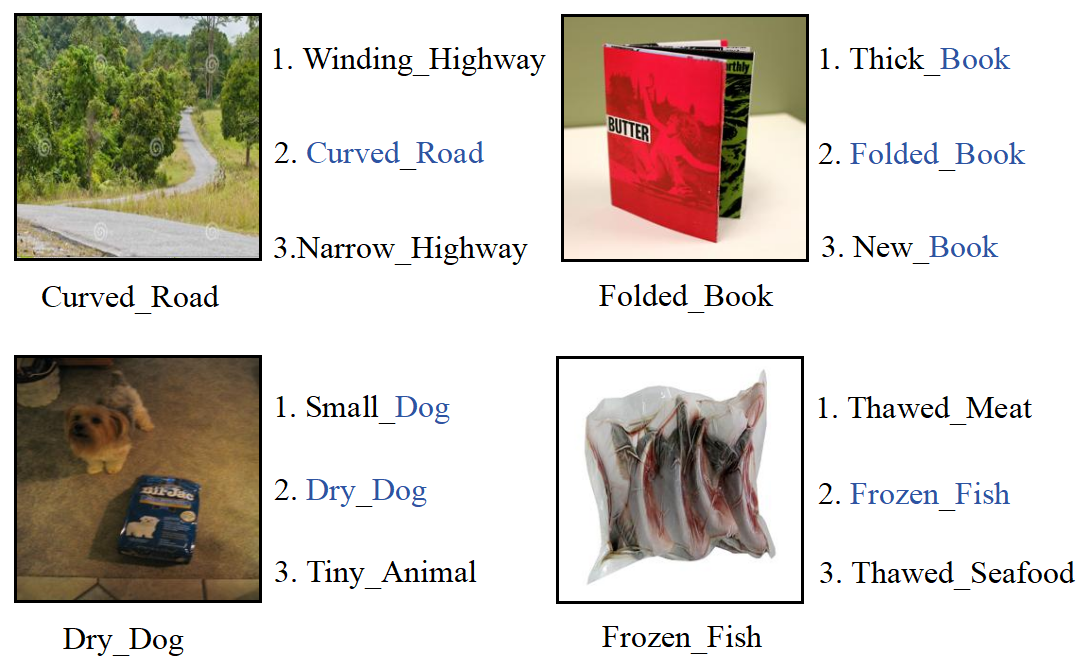}
    \vskip -0.1in
    \caption{For the failure qualitative results: Top-3 predictions for each sample are presented, and the correct ones are marked in blue.}
    \label{fig:visualization_failure}
\end{figure}

\subsection{Qualitative Analysis} 
\label{sec:qa}
In this section, we present some visualization results of CPF for both \textit{CW} (left) and \textit{OW} (right) settings in Fig.~\ref{fig:visualization}. Specifically, we report the top-3 prediction results for each sample, where the correct predictions are marked as blue. Our methods demonstrate stable attribute-object prediction under diversified challenging scenarios including a variety of outdoor scenes in MIT-States~\cite{isola2015discovering},  fine-grained attribute descriptions (various colors, material about shoes) in UT-Zappos50K~\cite{yu2014fine} as well as more complex C-GQA~\cite{naeem2021learning}. More qualitative results can be found in supplementary materials. 

\subsection{Failure Cases and Limitations} 
\label{sec:failure_discuss}
Though CPF improves zero-shot inference performance in CZSL, it occasionally demonstrates issues that are common to ambiguous scenes. In this section, we clarify the limitations of our proposed CPF and provide in-depth discussions. In particular, we present four examples of failure cases in MIT-States~\cite{isola2015discovering} (Fig.~\ref{fig:visualization_failure}). These failure cases can be attributed to two factors: i) there exists semantic ambiguity among class labels, such as ``highway'' \textit{vs} ``road'' and ``thick'' \textit{vs} ``folded'' in the first row; ii) The targets in images are visually confusing, such as the ``thawed meat'' is highly similar to the ``frozen fish'' in the bottom right. Therefore, we propose leveraging large language models to generate more discriminative textual descriptions for these semantically similar classes in the future. More qualitative discussion can be found in supplementary materials. 

\section{Conclusion}
\label{conclusion}
This paper introduces a Conditional Probability Framework (CPF) to model the interdependence between attributes and objects. We decompose composition probability into two components: object likelihood and conditional attribute likelihood. For object likelihood, we employ a text-enhanced object learning module that combines deep visual and textual embeddings to enhance object representations. For conditional attribute likelihood, we propose an object-guided attribute learning module that leverages text-enhanced object features and shallow visual embeddings to capture attribute-object relationships. By jointly optimizing both components, our method effectively models compositional dependencies and generalizes to unseen compositions. Extensive experiments on multiple CZSL benchmarks under both \textit{CW} and \textit{OW} settings demonstrate the superiority of our approach. The source code is publicly available at \href{https://github.com/Pieux0/CPF}{here}.


{
    \small
    \bibliographystyle{ieeenat_fullname}
    \bibliography{main}
}

\end{document}


\maketitle

This document offers theoretical justification and implementation of CPF, pseudo-code of CPF, additional experimental results, further qualitative analysis, comprehensive implementation details and extensive dataset information. The structure is organized as follows: 
\begin{itemize}
    \item \S\ref{sec:a0} Theoretical justification and implementation of CPF
    \item \S\ref{sec:a1} Pseudo-code of CPF
    \item \S\ref{sec:a2} More experimental results
    \item \S\ref{sec:a3} More qualitative analysis
    \item \S\ref{sec:a4} More implementation details
    \item \S\ref{sec:a5} Summary of data split statistics
\end{itemize}

\section{Theoretical Justification and Implementation of CPF}
\label{sec:a0}

The chain rule of probability universally decomposes any joint probability into a product of conditional and marginal probabilities~\cite{ho2020denoising}: $p(a,o|x)=p(a|o,x)p(o|x)$ (or symmetrically $p(a,o|x)=p(o|a,x)p(a|x)$). In CZSL, the choice to decompose $p(a,o|x)$ as $p(a|o,x)p(o|x)$ (rather than $p(a|x)p(o|x)$ or $p(o|a,x)p(a|x)$) is driven by semantic and contextual dependencies: (i) the independence assumption $p(a)p(o)$ fails to capture semantic binding~\cite{hu2024token,rassin2023linguistic}, whereas the conditional probability explicitly models plausible attribute-object relationships. (ii) The semantic asymmetry (established empirically by ~\citet{nagarajan2018attributes}) structurally favors $p(a|o,x)$ over $p(o|a,x)$, as objects act as semantic anchors that causally determine plausible attributes, whereas the inverse mapping $p(o|a,x)$ is ill-posed due to attribute-sharing across objects, violating injectivity for stable inference. 
In practical, we adopt the additive formulation to address the practical issue of ``probability vanishing'' inherent in multiplicative approaches.

\section{Pseudo-code of CPF}
\label{sec:a1}
Algorithm~\ref{alg:code} provides the pseudo-code of CPF.

\begin{algorithm}[H]
    \caption{Pseudo-code of CPF in a PyTorch-like style.}
    \label{alg:code}
    \definecolor{codeblue}{rgb}{0.25,0.5,0.5}
	\lstset{
		backgroundcolor=\color{white},
		basicstyle=\fontsize{7.8pt}{7.8pt}\ttfamily\selectfont,
		columns=fullflexible,
		breaklines=true,
		captionpos=b,
		escapeinside={(:}{:)},
		commentstyle=\fontsize{7.8pt}{7.8pt}\color{codeblue},
		keywordstyle=\fontsize{7.8pt}{7.8pt},
    }
   \begin{lstlisting}[language=python]
"""
# v_h_c: deep-level class feature (1 x D)
# V_h_p: deep-level patch feature (HW x D)
# v_l_c: shallow-level class feature (1 x D)
# V_h_p: shallow-level patch feature (HW X D)
# W_a: attribute textual embedding (M x D)
# W_o: object textual embedding (N X D)
# W_f_o: projection matrix (D x d)
# W_f_a: projection matrix (D x d)
# D: visual feature dimension
# d: textual embedding dimension
"""

#======= text-enhaced object learning =======#
# projecting deep-level class feature into the joint semantic space
v_h_c_d = v_h_c (:\color{codeorg}\textbf{@}:) W_f_o
# compute similarity score
score_1 =  (:\color{codeorg}\textbf{torch.matmul}:)(v_h_c_d, W_o.(:\color{codepro}{\textbf{transpose()}:)) /  (:\color{codeorg}\textbf{torch.sqrt}:)( (:\color{codeorg}\textbf{torch.tensor}:)(d))
# compute q_t with similarity score
q_t =  (:\color{codeorg}\textbf{F.softmax}:)(score_1, dim=-1)  (:\color{codeorg}\textbf{@}:) W_o
# projecting deep-level patch feature into the joint semantic space
V_h_p_d = V_h_p  (:\color{codeorg}\textbf{@}:) W_f_o
# compute attention score
score_2 =  (:\color{codeorg}\textbf{torch.matmul}:)(q_t, V_h_p.(:\color{codepro}{\textbf{transpose()}:)) /  (:\color{codeorg}\textbf{torch.sqrt}:)( (:\color{codeorg}\textbf{torch.tensor}:)(d))
# compute object feature
v_o = v_h_c +  (:\color{codeorg}\textbf{F.softmax}:)(score_2, dim=-1)  (:\color{codeorg}\textbf{@}:) V_h_p

#======= object-guided attribute learning =======#
# compute attention score
score_3 =  (:\color{codeorg}\textbf{torch.matmul}:)(v_o, V_l_p.(:\color{codepro}{\textbf{transpose()}:)) /  (:\color{codeorg}\textbf{torch.sqrt}:)( (:\color{codeorg}\textbf{torch.tensor}:)(D))
# compute attribute feature
v_a =  (:\color{codeorg}\textbf{F.softmax}:)(score_3, dim=-1)  (:\color{codeorg}\textbf{@}:) V_l_p

   \end{lstlisting}
\end{algorithm}

\section{More Experiment Results}
\label{sec:a2}
In this section, we present the ablation study on loss weight coefficients $\alpha_1$ and $\alpha_2$ on UT-Zappos50K~\cite{yu2014fine}. The results are shown in Table~\ref{tab:weights}. Additionally, we conduct ablation experiments on block choices on UT-Zappos50K~\cite{yu2014fine} to select the most suitable blocks as shallow-level visual embeddings. The corresponding results are presented in Table~\ref{tab:block_choice}. Moreover, as shown in the Table~\ref{tab:clip_based}, our CLIP-based CPF consistently outperforms other methods on MiT-States and UT-Zappos50K.

\begin{table}[ht]
  \setlength{\abovecaptionskip}{0cm}
	\setlength{\belowcaptionskip}{0cm}
    \caption{Ablation study on loss weight coefficients $\alpha_1$ and $\alpha_2$ on UT-Zappos50K~\cite{yu2014fine}.}
    \label{tab:weights}
    \centering
    \renewcommand{\arraystretch}{1.2}
    \resizebox{0.40\textwidth}{!}{
    \begin{tabular}{|c|c||cccc|}
    \thickhline
    \rowcolor{mygray}
    &  &  \multicolumn{4}{c|}{UT-Zappos50K} \\
    \rowcolor{mygray}
    $\alpha_1$ & $\alpha_2$  & AUC$\uparrow$ & HM$\uparrow$ & Seen$\uparrow$ & Unseen$\uparrow$ \\ 
    \hline
    0.0 & 0.0 & 28.5 & 45.1 & 56.4 & 60.9  \\
    0.3 & 0.7 & 39.2 & 53.3 & 65.3 & 71.3  \\
    0.4 & 0.6 & 39.3 & 53.4 & 64.3 & \textbf{72.5} \\
    0.5 & 0.5 & 40.1 & 55.3 & 65.6 & 69.4   \\
    0.6 & 0.4 & \textbf{41.4} & \textbf{55.7} & \textbf{66.4} & 71.1  \\
    0.7 & 0.3 & 40.0 & 54.5 & 66.0 & 69.3  \\
    \hline
    \end{tabular}}
\end{table}

\begin{table}[ht]
\setlength{\abovecaptionskip}{0cm}
\setlength{\belowcaptionskip}{0cm}
    \caption{Ablation study on block choices on UT-Zappos50K~\cite{yu2014fine}.}
    \vskip 0.15in
    \centering
    \renewcommand{\arraystretch}{1.2}
    \resizebox{0.45\textwidth}{!}{
    \begin{tabular}{|c|r||cccc|}
    \thickhline 
    \rowcolor{mygray}
    & & \multicolumn{4}{c|}{UT-Zappos50K} \\
    \rowcolor{mygray}
    Setting & Blocks & AUC$\uparrow$ & HM$\uparrow$ & Seen$\uparrow$ & Unseen$\uparrow$ \\ 
    \multirow{3}{*}{\textit{\makecell[c]{Close\\World}}} & (1,4,7) & 38.3 &52.4	&65.5 &71.1  \\
    & (2,5,8) & 39.2 &53.6	&65.3	&\textbf{71.7}  \\
    & (3,6,9) & \textbf{41.4} &	\textbf{55.7}	&\textbf{66.4}	&71.1 \\
    \hline
    \multirow{3}{*}{\textit{\makecell[c]{Open\\World}}} & (1,4,7) & 28.8	&45.6	&64.1	&52.3  \\
    & (2,5,8) & 29.0	&46.2	&\textbf{65.3}	&51.0  \\
    & (3,6,9) & \textbf{31.2}	&\textbf{47.6}	&64.6	&\textbf{56.1}  \\
    \thickhline
    \end{tabular}}
\label{tab:block_choice}
\end{table}

\begin{table}[h]
\setlength{\abovecaptionskip}{0cm}
\setlength{\belowcaptionskip}{0cm}
\caption{Results of CLIP-based CPF on MiT-States~\cite{isola2015discovering} and UT-Zappos50KK~\cite{yu2014fine}.}
    \vskip 0.15in
    \centering
    \renewcommand{\arraystretch}{1.2}
    \resizebox{0.40\textwidth}{!}{
    \footnotesize
    \begin{tabular}{|c|cc|cc|}
    \thickhline
    \rowcolor{mygray}
    & \multicolumn{2}{c|}{UT-Zappos50K} & \multicolumn{2}{c|}{MiT-States} \\
    \rowcolor{mygray}
     Method & AUC$\uparrow$ & HM$\uparrow$  & AUC$\uparrow$  & HM$\uparrow$  \\ \hline 
     CDS-CZSL~\cite{li2024context} & 39.5 &52.7 & 22.4 &39.2 \\
     Troika~\cite{huang2024troika} & 41.7 &54.6 & 22.1 &39.3 \\
     PLID~\cite{bao2024prompting} & 38.7 & 52.4 & 22.1 & 39.0 \\
     CAILA~\cite{zheng2024caila} & 44.1 &57.0 & 23.4 & 39.9 \\
     Ours & \textbf{45.2} & \textbf{57.6} &\underline{23.2} & \textbf{40.5} \\
    \thickhline
    \end{tabular}
    }
    \label{tab:clip_based}
\end{table}

\section{More Qualitative Analysis}
\label{sec:a3}
\begin{figure}[h]
    \centering
    \includegraphics[width=1\linewidth]{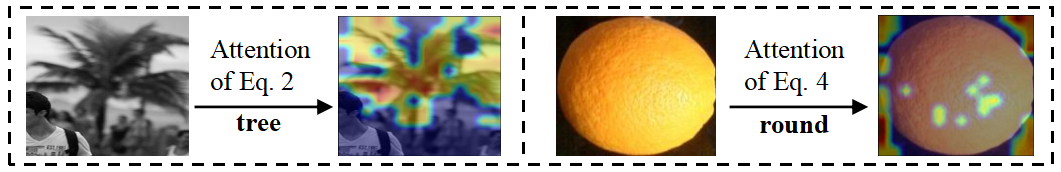}
    \vskip -0.1in
    \caption{Attention visualizations for Eq. \textcolor[RGB]{54,129,189}{2} and Eq. \textcolor[RGB]{54,129,189}{4}}
    \label{fig:figure_visualization}
\end{figure}

\begin{figure}[h]
    \centering
    \includegraphics[width=1\linewidth]{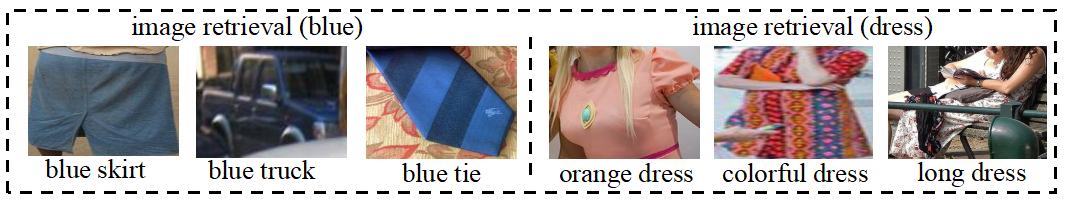}
    \caption{Qualitative results of image retrieval}
    \label{fig:image_retrieval}
\end{figure}

\begin{figure}[ht]
    \centering
    \includegraphics[width=1\linewidth]{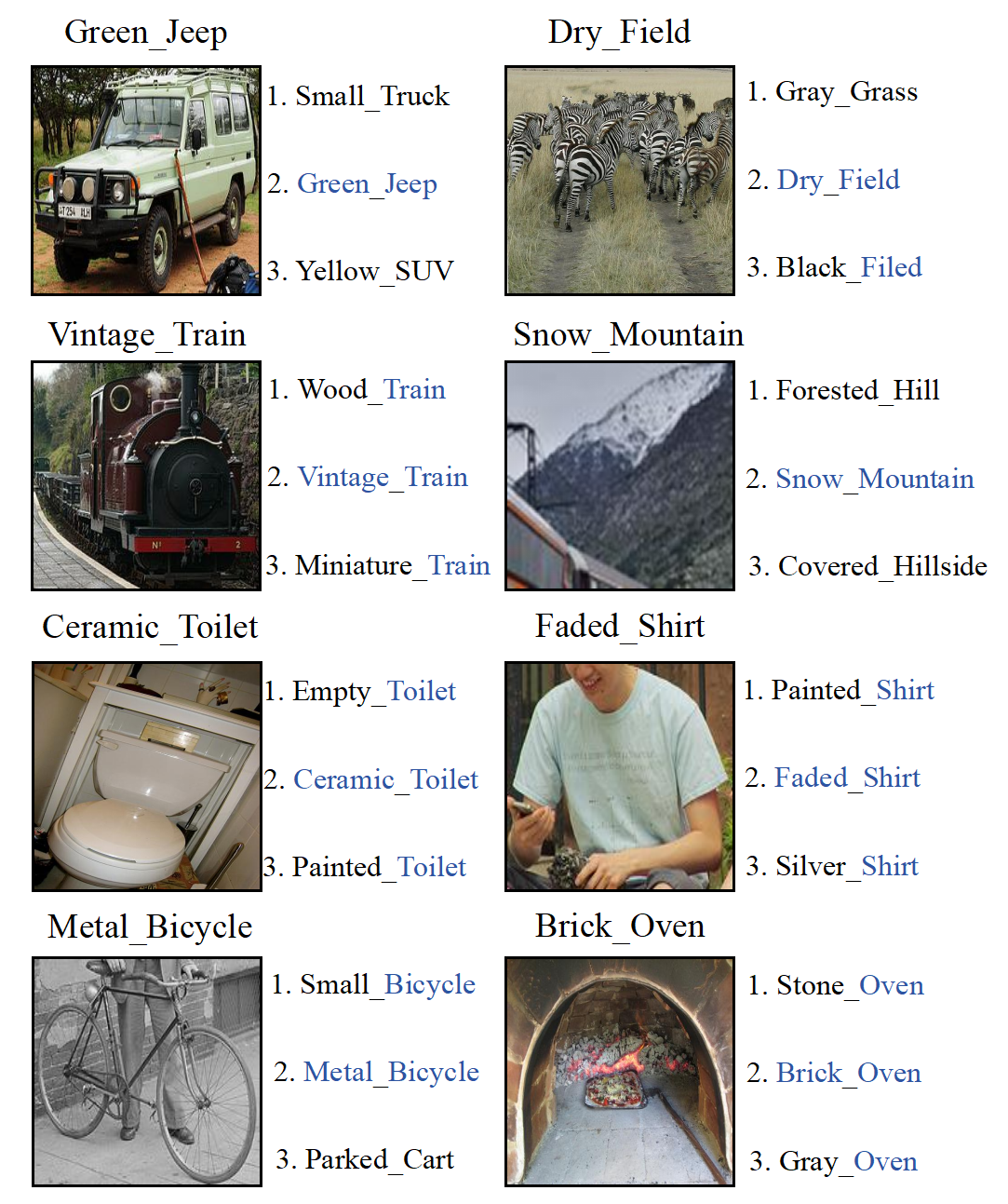}
    \vskip -0.1in
    \caption{More qualitative results of wrong predictions of instances in C-GQA~\cite{naeem2021learning}.}
    \label{fig:app_f_visualization}
\end{figure}

\begin{figure*}[ht]
    \centering
    \includegraphics[width=1\linewidth]{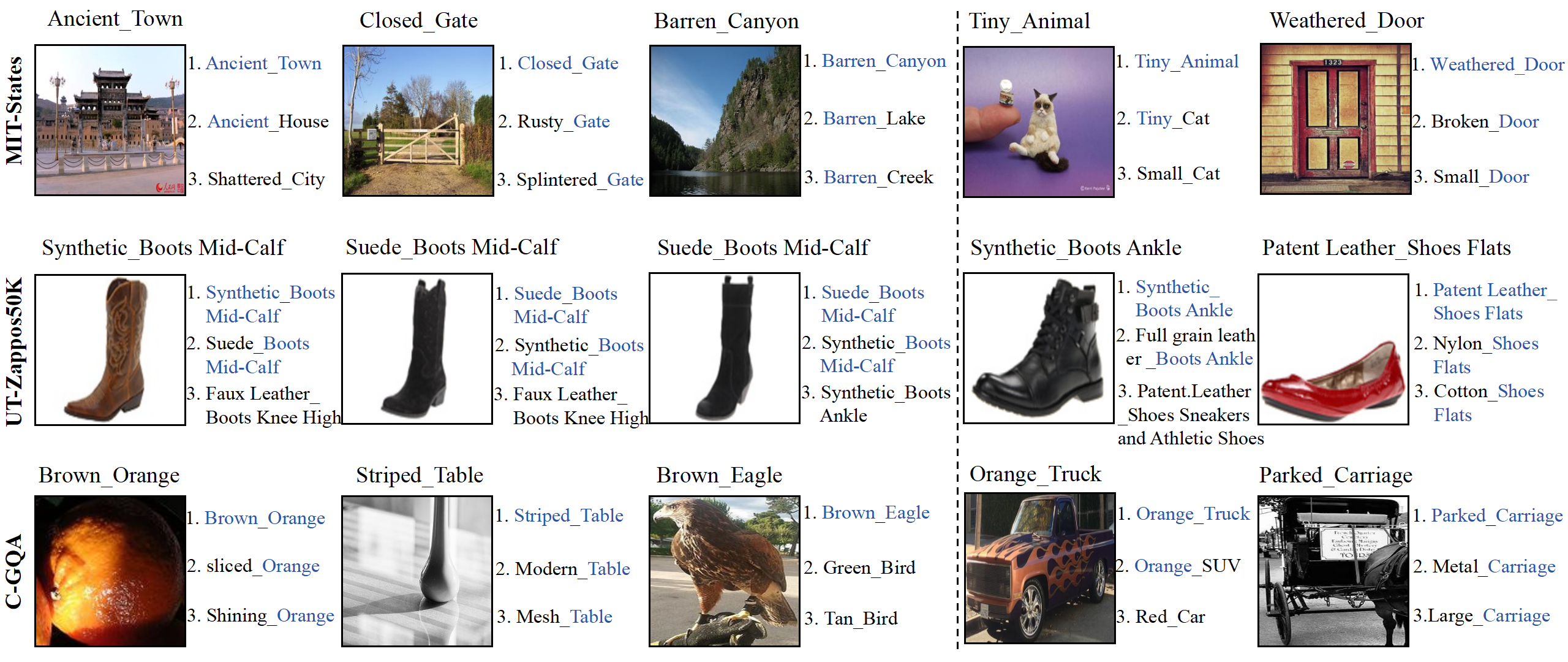}
    \caption{More qualitative results of UT-Zappos50K~\cite{yu2014fine}, MIT-States~\cite{isola2015discovering} and C-GQA~\cite{naeem2021learning}.}
    \label{fig:app_visualization}
\end{figure*}

We provide more qualitative results of UT-Zappos50K~\cite{yu2014fine}, MIT-States~\cite{isola2015discovering} and C-GQA~\cite{naeem2021learning} under \textit{CW} and \textit{OW} settings in Fig.~\ref{fig:app_visualization}. We show results for each dataset in each row. Images predicted under the \textit{CW} setting are shown in the first three columns and the rest of the columns show the instances under the \textit{OW} setting. In Fig.~\ref{fig:figure_visualization}, we provide attention visualization for Eq. \textcolor[RGB]{54,129,189}{2} and Eq. \textcolor[RGB]{54,129,189}{4}. In Fig.~\ref{fig:image_retrieval}, we illustrate qualitative results of image retrieval. In Fig.~\ref{fig:app_f_visualization}, we show additional wrong predictions of instances in  C-GQA~\cite{naeem2021learning}.

\section{More Implementations Details}
\label{sec:a4}
\begin{figure}[ht]
    \centering
    \includegraphics[width=0.99\linewidth]{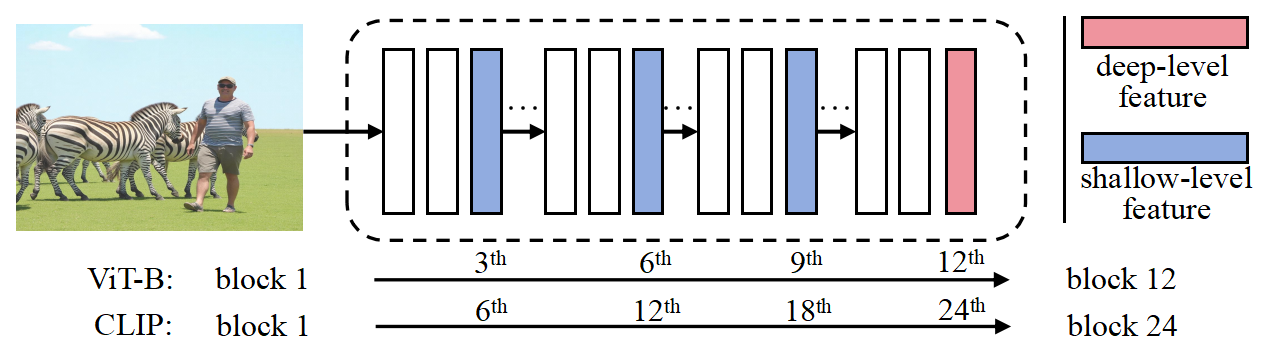}
    \caption{Implementation details of deep-level and shallow-level visual embeddings for both ViT-B and CLIP.}
    \label{fig:extract_details}
\end{figure}

We provide the implementation details of deep-level and shallow-level feature extraction in Fig.~\ref{fig:extract_details}.

\begin{table*}[ht]
      \setlength{\abovecaptionskip}{0cm}
	\setlength{\belowcaptionskip}{0cm}
    \caption{Summary of data split statistics.}
    \label{tab:data-splits}
    \centering
    \scalebox{0.90}{
    \begin{tabular}{|r||ccc|cc|cc|cc|}
        \thickhline
        \rowcolor{mygray}
         & \multicolumn{3}{c|}{Composition} & \multicolumn{2}{c|}{Train} & \multicolumn{2}{c|}{Val} & \multicolumn{2}{c|}{Test}  \\
         \rowcolor{mygray}
         Datasets & $|\mathcal{A}|$ & $|\mathcal{O}|$ & $|\mathcal{A}|\times|\mathcal{O}|$ & $|\mathcal{C}_{s}|$ & $|\mathcal{X}|$ & $|\mathcal{C}_{s}|$ / $|\mathcal{C}_{u}|$ & $|\mathcal{X}|$ & $|\mathcal{C}_{s}|$ / $|\mathcal{C}_{u}|$ & $|\mathcal{X}|$
         \\ 
         \hline
        UT-Zappos50K & 16 & 12 & 192 & 83 & 22998 & 15 / 15 & 3214 & 18 / 18 & 2914 \\
        MIT-States & 115 & 245 & 28175 & 1262 & 30338 & 300 / 300 & 10420 & 400 / 400 & 12995 \\
        C-GQA & 413 & 674 & 278362 & 5592 & 26920 & 1252 / 1040 & 7280 & 888 / 923 & 5098 \\
        \thickhline
    \end{tabular}}
\end{table*}

\section{Summary of Data Split Statistics}
\label{sec:a5}
Following previous work~\cite{hao2023learning,wang2023learning}, we provide the summary of data split statistics for UT-Zappos50K~\cite{yu2014fine}, MIT-States~\cite{isola2015discovering} and C-GQA~\cite{naeem2021learning} in Table~\ref{tab:data-splits}. $|\mathcal{A}|$ and $|\mathcal{O}|$ represent the numbers of attribute and object classes, respectively. $|\mathcal{C}_{s}|$ and $|\mathcal{C}_{u}|$ denote the numbers of seen and unseen composition categories, respectively. $|\mathcal{X}|$ indicates the numbers of images.

\clearpage

{
    \small
    \bibliographystyle{ieeenat_fullname}
    \bibliography{main}
}